\def\year{2021}\relax
\documentclass[letterpaper]{article} 
\usepackage{aaai21}  
\usepackage{times}  
\usepackage{helvet} 
\usepackage{courier}  
\usepackage[hyphens]{url}  
\usepackage{graphicx} 
\usepackage{natbib}  
\usepackage{caption} 
\usepackage{latexsym}
\usepackage{graphicx}
\usepackage{amsmath}
\usepackage{amsthm}
\usepackage{booktabs}
\usepackage{algorithm}
\usepackage{algorithmic}
\usepackage{multirow}
\usepackage{multicol}
\usepackage{arydshln}
\usepackage[T1]{fontenc}
\usepackage{tabu}  
\usepackage{multicol}  
\usepackage{multirow} 
\usepackage[switch]{lineno}
\usepackage{xspace}
\usepackage{amsfonts}
\title{SenSeNet: Neural Keyphrase Generation with Document Structure}

\title{SenSeNet: Neural Keyphrase Generation with Document Structure}
\author {

        Yichao Luo\textsuperscript{\rm 1},
        Zhengyan Li\textsuperscript{\rm 1},
        Bingning Wang\textsuperscript{\rm 2},
        Xiaoyu Xing\textsuperscript{\rm 1},
        Qi Zhang\textsuperscript{\rm 1}\thanks{Corresponding author},
        Xuanjing Huang\textsuperscript{\rm 1},
        \\
}
\affiliations {
    \textsuperscript{\rm 1} Shanghai Key Laboratory of Intelligent Information Processing\\
 School of Computer Science, Fudan University, Shanghai, China \\
    \textsuperscript{\rm 2} Sogou Inc., Beijing, China \\
    \{ycluo18,lizy19,xyxing18,qz,xjhuang\}@fudan.edu.cn,
    wangbingning@sogou-inc.com
}
\begin{document}
\maketitle

\begin{abstract}
Keyphrase Generation (KG) is the task of generating central topics from a given document or literary work, which captures the crucial information necessary to understand the content. Documents such as scientific literature contain rich meta-sentence information, which represents the logical-semantic structure of the documents. 
However, previous approaches ignore the constraints of document logical structure, and hence they mistakenly generate keyphrases from unimportant sentences.
To address this problem, we propose a new method called Sentence Selective Network (SenSeNet) to incorporate the meta-sentence inductive bias into KG. In SenSeNet, we use a straight-through estimator for end-to-end training and incorporate weak supervision in the training of the sentence selection module. Experimental results show that SenSeNet can consistently improve the performance of major KG models based on seq2seq framework, which demonstrate the effectiveness of capturing structural information and distinguishing the significance of sentences in KG task.
\end{abstract}

\section{Introduction}

\begin{figure}[t]
\centering
\includegraphics[width=1.0\columnwidth]{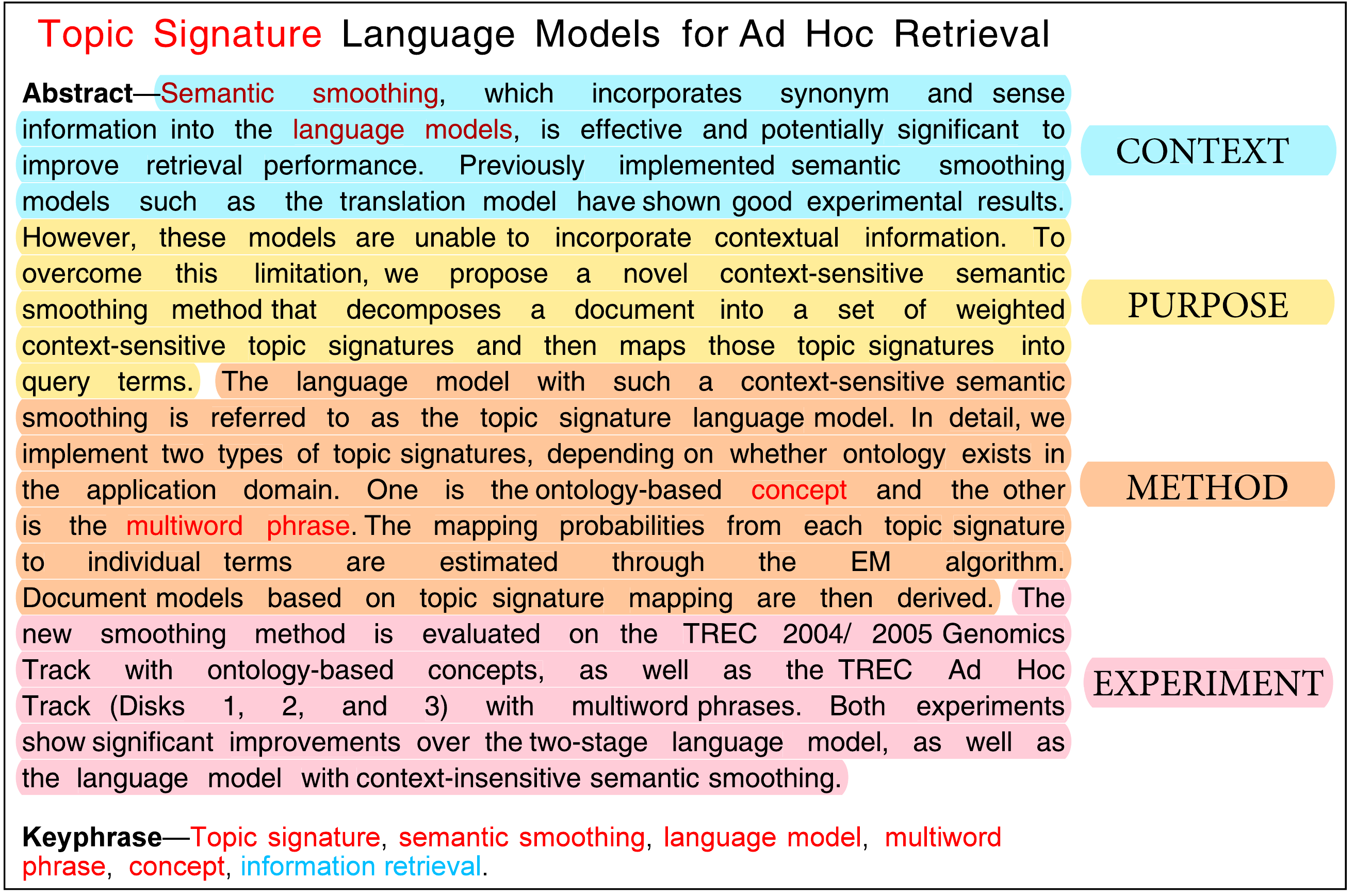}
\caption{An example of keyphrase generation. In this example, an abstract is categorized into four classes: context, purpose, method, and experiment. The red words represent the present keyphrase and The blue words represent the absent keyphrase.}
\label{fig:example}
\end{figure}

Keyphrase Generation (KG) is a traditional and challenging task in Natural Language Processing (NLP) which summarizes a given document and captures the central idea with keyphrases. Keyphrases allow one to conveniently understand the content of a document without reading the whole text. As KG provides a concise output and the core idea of the document, it is used for various downstream tasks, such as text categorizing~\cite{hulth2006study}, summarization~\cite{wang2013domain}, opinion mining~\cite{berend2011opinion}, etc. 

In KG, a keyphrase can be categorized into \textit{present} (red in Figure \ref{fig:example}) and \textit{absent} (blue in Figure \ref{fig:example}) depending on whether it exists in the original text. Early works focus on generating present keyphrase \cite{liu2010automatic,medelyan2009human}, and recently, 
many studies focus on producing both present and absent keyphrase. CopyRNN~\cite{meng2017deep} utilizes an attention-based sequence-to-sequence framework \cite{luong2015effective}, and incorporates a copy mechanism ~\cite{gu2016incorporating} to effectively generate rare words. CorrRNN ~\cite{chen2018keyphrase} leverages the Coverage~\cite{tu2016modeling} mechanism and review mechanism to incorporate the correlation among keyphrases. TG-Net ~\cite{chen2019guided} effectively exploits the title information to generate more consistent keyphrases. Based on sequence-to-sequence framework, \citet{chan2019neural} improves the performance through reinforcement learning with adaptive rewards, which encourages the model to generate succinct and accurate keyphrases.

While sequence-to-sequence based models are promising, they are insufficient to capture the logical structure of documents. For example, in the field of scientific literature, the keyphrases have a greater probability of being in the title and methodology. Take Figure \ref{fig:example} as an example, the generated keyphrases \textit{semantic smoothing} come from the \textsc{context}, and \textit{concept} come from \textsc{method}. It's obvious that the other sentences rarely related to key idea. Therefore, using hierarchical structure of the document to model the meta-sentence information could help to remove the irrelevant information, and focus our model on the central idea of the text. 


In this paper, we propose a novel model named SenSeNet to automatically capture the logical structure of documents and incorporate the meta-sentence inductive bias. 
First, we generate the hidden representation of each token with an encoder layer, such as a bi-directional Gated Recurrent Unit (GRU)~\cite{cho2014learning}. Then we use the sentence selection module to introduce inductive bias. In this module, a Convolutional Neural Networks (CNN) are applied to aggregate the word representations and obtain the sentence representations. After it, we get a binary output to determine whether the sentence tends to generate keyphrase. We translate the binary signal to an embedding represent and add into hidden representation from the encoder. Finally, we put the new representation into the decoder to generate finally keyphrases.

However, the output of the sentence selection module is binary which is discontinuous and blocks the gradient flow. To properly train the model, we use the straight-through estimator~\cite{bengio2013estimating} to detach the forward and backward process. In specific, we use the binary signal in forward-propagating and use the corresponding probability to compute the gradient approximately in back-propagation. In this way, the sentence selection module can be properly trained to induce important sentences by itself. Inspired by the characteristics of KG task, we design a reasonable signal to train the sentence selection module, which surprisingly boost the performance.

We evaluate our model on five public datasets for keyphrase generation. Experiments demonstrate that our model selects the significant sentences accurately and when applying our model to most KG methods, our model could achieve consistent improvements on both present and absent keyphrase, especially on absent keyphrase. We further analyze the predictions and propose a new concept:\textit{semi-present keyphrase}, in which all the words exist in the source text but are not the continuous subsequence of the source text (specific example is shown in table \ref{tab:casepsa}). We find that our model predicts semi-present keyphrase more accurately, which causes the better results on absent keyphrase. And we find that our model still performs well when the number of sentences increases, which proves the effectiveness of removing noise sentences. In addition, we also conduct some experiments to demonstrate that our model can be properly adapted to most encoder-decoder architectures.

\begin{figure*}[ht]
\centering
\includegraphics[width=0.95\linewidth]{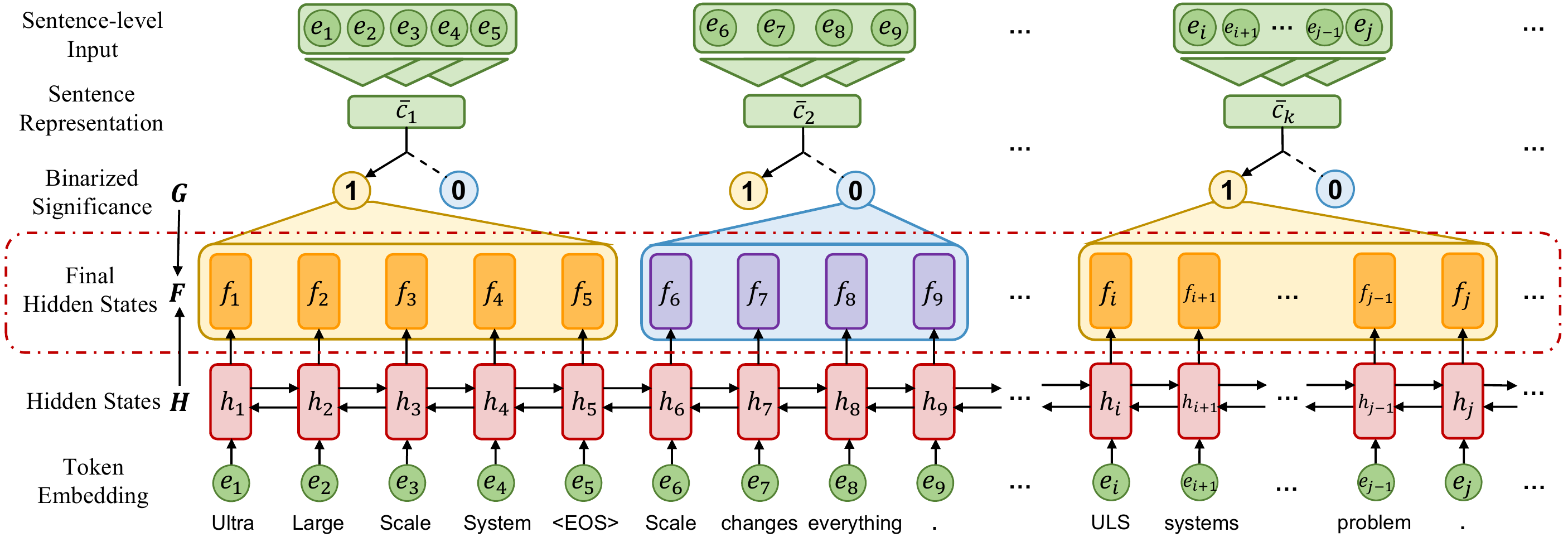}
\caption{The sentence selective encoder module (best viewed in color). The sentence selection module is above the red dot box, and the traditional encoder is below. In the box is the final state for encoder combining above two modules. Triangles represent CNN. ``1'' represents ``significant'' and ``0'' represents ``irrelevant''. The yellow and blue boxes represent the embedding information of ``1'' and ``0'' respectively. The orange and purple rectangles represent hidden states that merge into the embedding of ``1'' and ``0'' respectively.} 
\label{fig:model}
\end{figure*}

We summarize our contribution as follows:
\begin{itemize}
\item We propose a novel keyphrase generation method that can automatically learn the document structure, capture meta-sentence information, and guide the keyphrase generation with sentence selection information. Experiments demonstrate the generality of our model which can be adapted to most encoder-decoder architectures.
\item We use the straight-through estimator to train the model with an end-to-end manner. And we also propose reasonable weakly supervised information for training. 
\item We achieve improvements on both present and absent keyphrases in five KG datasets. And we propose a new concept: semi-present keyphrase to show the effectiveness of our model on absent keyphrase. A large number of analytical experiments have proved the effectiveness of our model.
\end{itemize}

\section{Related Work}
\paragraph{Keyphrase Generation}
Traditional keyphrase production methods directly extract significant spans as keyphrases through candidates extracting ~\cite{wang2016ptr,hulth2003improved} and ranking \cite{nguyen2007keyphrase,mihalcea2004textrank}. However, these traditional methods can only extract keyphrases that appear in the text with a sequential order. To produce absent keyphrases, which  do not appear as any continuous subsequence of text, sequence-to-sequence based models become a popular trend. 
Techniques include 
CopyRNN ~\cite{meng2017deep}, CorrRNN ~\cite{chen2018keyphrase} with coverage mechaism~\cite{tu2016modeling} and reviewer mechanism, TG-Net~\cite{chen2019guided}, and reinforcement learning~\cite{chan2019neural}. However, these works do not consider the meta-sentence information or the logical structure of the document, which provides important clues for generating keyphrases.

\paragraph{Incorporating Document Structure} Documents such as science literatures and news articles are well structured. The hierarchical structure of the documents are used in various NLP tasks to boost the performance. \citet{yang2016hierarchical} incorporate the knowledge of the hierarchical structure to text classification model. \citet{alzahrani2012using} utilizes the structural information to detect significant plagiarism problems. \citet{frermann2019inducing} induce latent document structure for generating aspect-based summarization. And in Keyphrase Extraction area, \citet{nguyen2010wingnus} propose to add document structure to hand-crafted features to help identify candidates, and \citet{chen2019guided} treat the document title as a dominant role to the overall document and let it guide the generation process. And in our work, we consider the logical structure of the whole documents (e.g. title, context, method), and propose a sentence selection module to determine which part of the document is important. However it is difficult for the module to learn the binary indicator from scratch, and hence, we adopt weakly supervised learning in our training process.

\newcommand{\tabincell}[2]{
}  
\begin{table}[!t]  
  \centering  
  \scriptsize 

  \begin{tabular}{ll}  
    \\[-2mm]  
    \toprule 
    SOURCE     &   A B C D E F G H I J\\  
    \midrule  
    \vspace{1mm}\\[-3mm]  
    \vspace{1mm}  
    present        &  "A B C", "E F G", "H I J", etc.\\  
     \vspace{1mm}  
    semi-present          &  "A B D", "B C A", "A D H", etc. \\  
     \vspace{1mm}  
    absent  &   "A B D", "B C A", "X Y Z", etc. \\  
    \bottomrule 
  \end{tabular}  
  \caption{Case of three kinds of keyphrase, A letter stands for a word. Semi-present keyphrase is that all words in this keyphrase exist in a particular sentence. }    \label{tab:casepsa}  
\end{table}

\section{Methodology}
\subsection{Problem Definition}
We introduce the formal representation of the problem for keyphrase generation as follow. Give a source document $\mathbf{x}$, our objective is to produce several ground-truth keyphrases $\mathcal{Y}=\{\mathbf{y}_1,...,\mathbf{y}_{|\mathcal{Y}|}\}$. Both source document $\mathbf{x} = (x^{1},...,x^{|\mathbf{x}|})$ and target keyphrases $\mathbf{y}_{i} = (y^{1}_{i},...,y^{|\mathbf{y}_{i}|}_{i})$ are word sequences, where $|\mathbf{x}|$ and $|\mathbf{y}_{i}|$
represent the length of source sequence $\mathbf{x}$ and the $i$-th keyphrase sequence $\mathbf{y}_{i}$, respectively. 
The overall architecture of SenSeNet is shown in Figure \ref{fig:model}. In the following, we will introduce each part of SenSeNet in detail. 

\subsection{Sentence Selective Encoder}
The sentence selective encoder aims to identify which sentence of the document is significant in keyphrase generation process. In this module, we firstly lookup embedding vector from embedding matrix for each word to map the discrete language symbols $\mathbf{x} = (x_1,x_2,...,x_T)$ to continuous embedding vectors $\mathbf{e} = (e_1,e_2,...,e_T)$. Then, we adopt an encode-layer (e.g. bi-directional Gated Recurrent Unit~\cite{cho2014learning} or Transformer~\cite{vaswani2017attention}) to obtain the hidden representations $\mathbf{H}$ :
\begin{align}
\mathbf{H} = \mathrm{Encode}(\mathbf{e})
\end{align}

After extracting features for words in the document, our next step is to obtain the sentence representations. In a document, the $i$-th sentence $\mathbf{S}_i$ of length $|\mathbf{S}_i|$ is represented as:
\begin{equation}
\mathbf{S}_{i} = e_{\pi_{i}} \oplus e_{\pi_{i}+1} \oplus...\oplus e_{\pi_{i}+|\mathbf{S}_i |-1}
\end{equation}
where $\oplus$ is the concatenation operator, and $\pi_{i}$ represents the beginning position of first word of the sentence $i$. Then, we use the Convolutional Neural Network (CNN) to obtain the representation for each sentence of the document\cite{kim2014convolutional}. The feature of sentence $\mathbf{S}_i$ is compressed by a filter $\mathbf{W}$ with the window size $k$. For example, the new feature the window of words from position $\pi_{i}+(j-1)k$ to $\pi_{i}+jk-1$ is represented as: 
\begin{align}
    c_{i}^j &= \sigma\left(\mathbf{W}\cdot\mathbf{e}_{\pi_{i}+(j-1)k:\pi_{i}+jk-1}+b\right)
\end{align}
where $b$ is a bias term and $\sigma$ is a non-linear function. The filter $\mathbf{W}$ is applied to each window of words wth size k in the document and produce a feature map:
\begin{align}
\mathbf{c}_i &= \left[c^1,c^2,...,c^{|S_i |-k+1}\right]
\end{align}
Then we apply a max-pooling operation on the feature map and take the maximum value $\bar{c}_i$ = max$\{\mathbf{c}_i\}$ as representation for sentence $\mathbf{S_i}$.

To predict whether the sentences of the document are inclined to generate keyphrases, we map the sentence representation to a binary label:
 \begin{align}
    \mathbf{m}_i & = \mathrm{MLP}(\bar{c}_i) \\
    \eta_i &  = \mathrm{sigmoid}\left(\mathbf{m}_{i}^T\mathbf{w}_{m_i}\right) \\
    z_i & =
    \begin{cases} 
        1,  & \eta_i > 0.5 \\
        0, & \eta_i \leq 0.5
    \end{cases}
    \label{eq7}
\end{align}
where $\mathrm{MLP}$ is the multi-layer perception, $\mathrm{sigmoid}$ is the sigmoid activiation function, $\mathbf{w}_{m_i}$ is a weight vector, and $z_i$ is a binary gate to determine whether the sentence $i$ is significant ($1$) or not ($0$).

In order to convey the importance of the sentence to all the words of that sentence, we convert the sentence-level binarized significance vector $\mathbf{z}$ to the token-level vector $\mathbf{g}$. Then we use an embedding matrix $\mathbf{D}$ $\in {\mathcal{R}^{1\times 2d}}$ to map the binary sequence $\mathbf{g}$ of length $T$ to the continuous representation $ \mathbf{G}$. And the final state $\mathbf{F}$ of the encoder is represented as the sum of $\mathbf{H}$ and $ \mathbf{G}$, which is denoted as:
\begin{align}
    \mathbf{G} = \mathbf{g}^{T} * \mathbf{D} \\
    \mathbf{F} = \mathbf{H} + \mathbf{G}  
\end{align}
where $d$ is hidden size.

\subsection{Decoder}
For the decoder, the context vector $\mathbf{u}$ is computed as a weighted sum of final hidden representation $\mathbf{F}$ by attention mechanism~\cite{DBLP:journals/corr/BahdanauCB14}, which can represent the whole source text information at step $t$:
\begin{equation}
    \mathbf{u}_t = \sum_{j=1}^{T}{\alpha_{tj}\mathbf{F}_j}
 \end{equation}
where $\alpha_{tj}$ represents the correlation between the source sequence at position $j$ and the output of the decoder at position $t$.

Utilizing the context vector $\mathbf{u}_t$, the decoder can apply the traditional language model to generate the word sequence step by step:
\begin{align}
    \mathbf{s}_t = \mathrm{Decode}&(y_{t-1}, \mathbf{s}_{t-1}, \mathbf{u}_t)\\
    p_{g}\left(y_t|y_{<t}, \mathbf{F}\right) &= \mathrm{Softmax}(y_{t-1}, \mathbf{s}_t, \mathbf{u}_t)
\end{align}
where $s_t$ represents the hidden state of the decoder at step $t$,  $\mathrm{Softmax}$ is the softmax function which calculates the probability distribution of all the words in vocabulary, and $y_t$ represents the prediction at time step $t$.

Due to  out-of-vocabulary (OOV) problem, the above model cannot generate rare words. Thus we introduce the copy mechanism~\cite{gu2016incorporating} into the encoder-decoder framework to produce the OOV words from the source text directly. The probability of producing a token contains two parts---generate and copy:
\begin{align}
    p\left(y_t|y_{<t}, \mathbf{F}\right) &= p_{g}\left(y_t|y_{<t}, \mathbf{F}\right) + p_{c}\left(y_t|y_{<t}, \mathbf{F}\right)\\
    p_{c}\left(y_t|y_{<t}, \mathbf{F}\right) &= \frac{1}{Z} \sum_{j:x_{j}=y_{t}}{e^{\omega(x_j)}}, y_t \in \chi\\
    \omega(x_j) &= \sigma\left(\mathbf{F}_{j}^{T}\mathbf{W}_c\right)\mathbf{s}_t
\end{align}
where $\chi$ represents the set of all rare words in source document $\mathbf{x}$, $\mathbf{W}_c$ is a learnable matrix and $Z$ is used for normalization.

\begin{table*}[tp]  
	\centering\small
	\scalebox{0.80}{
	\begin{tabular}{l|c|c|c|c|c|c|c|c|c|c}
        \toprule
		\multirow{2}{*}{\textbf{Model}} & \multicolumn{2}{c|}{\textbf{Inspec}} &  \multicolumn{2}{c|}{\textbf{NUS}} & \multicolumn{2}{c|}{\textbf{Krapivin}} & \multicolumn{2}{c|}{\textbf{SemEval}} & \multicolumn{2}{c}{\textbf{KP20k}} \\ 
		& \multicolumn{1}{c}{$F_{1}@M$} &  \multicolumn{1}{c|}{$F_{1}@5$} & \multicolumn{1}{c}{$F_{1}@M$} &  \multicolumn{1}{c|}{$F_{1}@5$} &\multicolumn{1}{c}{$F_{1}@M$} &  \multicolumn{1}{c|}{$F_{1}@5$} &\multicolumn{1}{c}{$F_{1}@M$} &  \multicolumn{1}{c|}{$F_{1}@5$} &\multicolumn{1}{c}{$F_{1}@M$} &  \multicolumn{1}{c}{$F_{1}@5$} \\
        \midrule
		catSeq & 0.262 & 0.225 & 0.397 & 0.323 & 0.354 & 0.269 & 0.283 & 0.242 & 0.367 & 0.291 \\
		catSeqD  & 0.263 & 0.219 & 0.394 & 0.321 & 0.349 & 0.264 & 0.274 & 0.233 & 0.363 & 0.285 \\
		catSeqCorr  & 0.269 & 0.227 & 0.390 & 0.319 & 0.349 & 0.265 & 0.290 & 0.246 & 0.365 & 0.289 \\
		catSeqTG  & 0.270 & 0.229 & 0.393 & 0.325 & \textbf{0.366} & \textbf{0.282} & 0.290 & 0.246 & 0.366 & 0.292 \\
        \midrule
		SenSeNet  & \textbf{0.284} & \textbf{0.242} & \textbf{0.403} & \textbf{0.348} & 0.354 & 0.279 & \textbf{0.299} & \textbf{0.255} & \textbf{0.370} & \textbf{0.296} \\
		\midrule
		catSeq+RL  & 0.300 & 0.250 & 0.426 & 0.364 & 0.362 & 0.287 & 0.327 & 0.285 & 0.383 & 0.310 \\
		catSeqD+RL  & 0.292 & 0.242 & 0.419 & 0.353 & 0.360 & 0.282 & 0.316 & 0.272 & 0.379 & 0.305 \\
	    catSeqCorr+RL  & 0.291 & 0.240 & 0.414 & 0.349 & 0.369 & 0.286 & 0.322 & 0.278 & 0.382 & 0.308 \\
		catSeqTG+RL  & 0.301 & 0.253 & 0.433 & \textbf{0.375} & 0.369 & 0.300 & 0.329 & 0.287 & 0.386 &\textbf{0.321} \\
		\midrule
		SenSeNet+RL  & \textbf{0.311} & \textbf{0.278} & \textbf{0.440} & 0.374 & \textbf{0.375} & \textbf{0.301} & \textbf{0.340} & \textbf{0.296} & \textbf{0.389} & 0.320 \\
		\bottomrule
	\end{tabular} 
	}
	\caption{Result of present keyphrase prediction on five datasets. Suffix ``+RL'' denotes that a model is trained by reinforcement learning.} 
	\label{tab:present}
\end{table*}

\subsection{Training}
Finally, to train the model, we minimize the negative log likelihood loss:
\begin{equation}
    \mathcal{L}_{MLE} = -\sum{\mathrm{log}P\left(y_t|y_{<t}, \mathbf{F}\right)}
\end{equation}

However, in the encoder module, we generate the value of the binary gates from sentence representation Eq. \ref{eq7} , which causes the model to be discontinuous and blocks the gradient flow. To address this problem, a general approach is to use the policy gradient to sample an action from language model $p(y_t|y_{<t}, x)$, then the environment gives a reward to guide the generator in the next step. However, policy gradient has the well-known problem of high cost and variance, , which makes model difficult to train.

In this work, inspired by previous studies of training discontinuous neural networks~\cite{bengio2013estimating,courbariaux2016binarized}. We estimate the gradients for binary representation by the straight-through estimator. For a particular parameter $\theta$ in encoder, we estimate the gradient as follow:
\begin{align}
    \label{eq:16}
    &\frac{\mathrm{d}\mathbb{E} \left [\sum_{t}{\log{p}(y_{t}|y_{<t}, \mathbf{F})} \right ]}{\partial \theta} \nonumber\\
    =&\frac{\mathrm{d} \mathbb{E} \left [\sum_{t}{\log{p}(y_{t}|y_{<t}, \mathbf{F})} \right ]}{\mathrm{d} z} \frac{\mathrm{d} z}{\mathrm{d} \eta} \frac{\mathrm{d} \eta}{\partial  \theta} \\
    \approx&\frac{\mathrm{d}\mathbb{E}\left [\sum_{t}{\log{p}(y_{t}|y_{<t}, \mathbf{F})}\right ]}{\mathrm{d} z} \frac{\mathrm{d} z}{\mathrm{d} \eta}{\partial \theta}\nonumber
\end{align}

We estimate the gradient of hard variable $z$ using a smooth and continuous function $\eta$. Although it is a biased estimation, it can efficiently estimate the gradient of the discrete variables.

\subsection{Weakly Supervised Training}
We input the source document into the proposed model without introducing any structural information. Thus, in the training process, the sentence-selector should be trained by the model itself, which leads to the slow convergence and unstable training of the sentence-selector. To solve the problem, we apply a weakly supervised signal for each sentence. From previous work \cite{meng2017deep, chan2019neural}, We observe that most correct predictions come from present keyphrase, thus
based on this task characteristic, we propose a weakly-supervised label $a_i \in \{0,1\}$ for each sentence, which denotes whether the $i$-th sentence is significant. The type of weakly supervision is inaccurate supervision. We consider the sentences contain the present keyphrase and semi-present keyphrase are significant (1), otherwise irrelevant (0). Then we add the additional Binary Cross-Entropy loss to the supervised objective:
\begin{equation}
    \mathcal{L}_{BCE} = -\sum_{i}{a_{i}\log{\eta_i}} + (1 - a_i)\log(1-\eta_i)
\end{equation}
Thus the final supervised loss functions is to optimize:
\begin{equation}
    \mathcal{L} = \mathcal{L}_{MLE} + \lambda\mathcal{L}_{BCE}
\end{equation}
where $\lambda$ is a hyper-parameter.

\section{Experiment}
\subsection{Dataset}
We evaluate our model on five public scientific KG dataset:
\begin{itemize}
    \item \textbf{Inspec}~\cite{hulth2003improved} is a dataset that provides the abstract of 2,000 articles. The dataset are collected from journal papers and the disciplines \textit{Computers and Control}, \textit{Information Technology}.
    \item \textbf{NUS}~\cite{nguyen2007keyphrase} is a dataset that produce 211 articles. The dataset are from \textit{Google SOAP API.}
    \item \textbf{Krapivin}~\cite{krapivin2009large} is a dataset that provides 2304 pairs of source text(full paper) and author-assigned keyphrases. The dataset are collected from computer science domain, which is published by \textit{ACM}. 
    \item \textbf{SemEval-2010}~\cite{kim2010semeval} is a dataset that provides 288 articles collected from \textit{Semeval-2010 task}. 
    \item \textbf{KP20k}~\cite{meng2017deep} is a large dataset with more than 50,000 pairs of source text (title and abstract) and keyphrases. We take 20,000 pairs as validation set, 20,000 pairs as testing set, and the rest as training set. 
\end{itemize}

\subsection{Implementation Details}
Following \citet{chan2019neural}, we replace all the digits with token <\textit{digit}>, separate present keyphrases and absent keyphrases by token <\textit{peos}> and sorted present keyphrase. We use greedy search during testing and also remove all the duplicated keyphrases from the predictions. We select the top $50000$ words as our vocabulary, set the encoder hidden size to 300, 
and we initialize the model parameters using a uniform distribution range of $[-0.1, 0.1]$.

\begin{table*}[tp]  
	\centering\small
	\scalebox{0.8}{
	\begin{tabular}{l|c|c|c|c|c|c|c|c|c|c}
        \toprule
		\multirow{2}{*}{\textbf{Model}} & \multicolumn{2}{c|}{\textbf{Inspec}} &  \multicolumn{2}{c|}{\textbf{NUS}} & \multicolumn{2}{c|}{\textbf{Krapivin}} & \multicolumn{2}{c|}{\textbf{SemEval}} & \multicolumn{2}{c}{\textbf{KP20k}} \\ 
		& \multicolumn{1}{c}{$F_{1}@M$} &  \multicolumn{1}{c|}{$F_{1}@5$} & \multicolumn{1}{c}{$F_{1}@M$} &  \multicolumn{1}{c|}{$F_{1}@5$} &\multicolumn{1}{c}{$F_{1}@M$} &  \multicolumn{1}{c|}{$F_{1}@5$} &\multicolumn{1}{c}{$F_{1}@M$} &  \multicolumn{1}{c|}{$F_{1}@5$} &\multicolumn{1}{c}{$F_{1}@M$} &  \multicolumn{1}{c}{$F_{1}@5$} \\
        \midrule
		catSeq & 0.008 & 0.004 & 0.028 & 0.016 & 0.036 & 0.018 & 0.028 & 0.020 & 0.032 & 0.015 \\
		catSeqD  & 0.011 & 0.007 & 0.024 & 0.014 & 0.037 & 0.018 & 0.024 & 0.014 & 0.031 & 0.015 \\
		catSeqCorr  & 0.009 & 0.005 & 0.024 & 0.014 & 0.038 & 0.020 & 0.026 & 0.018 & 0.032 & 0.015 \\
		catSeqTG  & 0.011 & 0.005 & 0.018 & 0.011 & 0.034 & 0.018 & 0.027 & 0.019 & 0.032 & 0.015 \\
        \midrule
		SenSeNet  & \textbf{0.018} & \textbf{0.010} & \textbf{0.032} & \textbf{0.018} & \textbf{0.046} & \textbf{0.024} & \textbf{0.032} & \textbf{0.024} & \textbf{0.036} & \textbf{0.017} \\
        \midrule
		catSeq+RL  & 0.017 & 0.009 & 0.031 & 0.019 & 0.046 & 0.026 & 0.027 & 0.018 & 0.047 & 0.024 \\
		catSeqD+RL  & \textbf{0.021} & 0.010 & 0.037 & 0.022 & 0.048 & 0.026 & 0.030 & 0.021 & 0.046 & 0.023 \\
	    catSeqCorr+RL  & 0.020 & 0.010 & 0.037 & 0.022 & 0.040 & 0.022 & 0.031 & 0.021 & 0.045 & 0.022 \\
		catSeqTG+RL  & \textbf{0.021} & 0.012 & 0.031 & 0.019 & \textbf{0.053} & 0.030 & 0.030 & 0.021 & 0.050 & 0.027 \\
        \midrule
		SenSeNet+RL  & \textbf{0.021} & \textbf{0.013} & \textbf{0.039} & \textbf{0.026} & 0.048 & \textbf{0.031} & \textbf{0.033} & \textbf{0.024} & \textbf{0.055} & \textbf{0.030} \\
        \bottomrule
	\end{tabular}
	}
	\caption{Results of absent keyphrase prediction on five datasets. Suffix ``+RL'' denotes that a model is trained by reinforcement learning.} 
	\label{tab:absent}
\end{table*}

For the unique details of our model, we set the size of binary embedding to $300$ which is the same as the encoder hidden size. We divide sentences by ``.'' except for which in ``e.g.'' or ``i.e.''. The kernel size in CNN is set to $\{1, 3, 5\}$ and channel is set to $100$. The coefficient $\lambda$ of $\mathcal{L}_{BCE}$ is set to $0.08$, which is tuned in test data. 
\subsection{Baseline Models and Evaluation Metric}
In this paper, we only consider the model based on sequence-to-sequence framework. We select four previous models as our baseline models, catSeq ~\cite{yuan2018one} with a copy mechanism, catSeqD~\cite{yuan2018one} with a dynamic variable which denotes the length of predicting keyphrases, catSeqCorr~\cite{chen2018keyphrase} with coverage and reviewer mechanism, and catSeqTG~\cite{chen2019guided} with title-guide information. We refer to ~\citet{chan2019neural} for the name of the above models. All these baselines are based on catSeq. And for fair comparison, we add our proposed model SenSeNet on catSeq without other additional mechanisms.  According to ~\citet{chan2019neural}, reinforcement learning can empower keyphrase generation model by a large margin. Therefore we apply reinforcement learning in our SenSeNet model, and compare its performance with other baseline models using reinforcement learning. In addition, to prove the generalization of our model, we explore the influence of different encoder 
and decoder structures (e.g. Transformer or LSTM).

Most previous work~\cite{meng2017deep,chen2018keyphrase,chen2019guided} cut the fixed length predicting keyphrases to calculate metrics such as $F_{1}@5$ and $F_{1}@10$. To evaluate keyphrases with variable number, ~\citet{yuan2018one} proposes a new evaluation metric, $F_{1}@M$, which computes $F_1$ by comparing all predicted keyphrases  with the ground-truth.

\begin{table}[tp]  
	\centering\small
	\scalebox{0.8}{
	\begin{tabular}{l|c|c|c|c}
        \toprule
		\multirow{2}{*}{\textbf{Msodel}} & \multicolumn{2}{c|}{\textbf{Semi-present}} &
		\multicolumn{2}{c}{\textbf{Absent w/o}}
		\\ 
		& \multicolumn{1}{c}{\#count} &  \multicolumn{1}{c|}{Recall} & \multicolumn{1}{c}{\#count} &  \multicolumn{1}{c}{Recall}  \\
        \midrule
		catSeq & 398 & 0.053 & 637 & 0.020 \\
		catSeqD  & 384 & 0.051 & 674 & 0.021 \\
		catSeqCorr  & 408 & 0.054 & 676 & 0.021 \\
		catSeqTG  & 390 & 0.052 & 681 & 0.021 \\
        \midrule
		SenSeNet  & \textbf{440} & \textbf{0.059} & \textbf{763} & \textbf{0.024} \\
        \bottomrule
	\end{tabular}
	}
	\caption{Result of semi-present keyphrase and absent w/o (without semi-present) keyphrase prediction on KP20k dataset. }
	\label{tab:semi-present}
\end{table}

We evaluate the performance of models on both present and absent keyphrase by $F_1@5$ and $F_1@M$. The definition of $F_1@5$ is the same as ~\cite{chan2019neural}. If the model generates less than five predictions, we will add wrong answers to the prediction until it reaches five predictions. It is different from the methods based on the beam search because they can generate a large number of keyphrases (like beam size = 50) and cut first 5 or 10.

\section{Result and Analysis}
\subsection{Present and Absent Keyphrase Prediction}
In this section, we evaluate the performance of models on present and absent keyphrase predictions, respectively. Table \ref{tab:present} shows the predicted results for all the models on predicting present keyphrases. From the results, we have the following observations. \textit{\textbf{First}}, on most datasets our model outperforms all the baselines on both $F_{1}@5$ and $F_{1}@M$, which demonstrates the validity of introducing document structure into the KG neuron model. \textit{\textbf{Second}}, on \textit{Krapivin} dataset, carSeqTG performs better than us. From the analysis of keyphrases distribution in \textit{Krapivin}, we observe 35.5\% keyphrases exist in the documents' titles, a much higher percentage than other datasets. And catSeqTG explicitly models title feature, which makes it perform better on the Krapivin dataset. \textit{\textbf{Third}}, our proposed model SenSeNet performs well on present keyphrase prediction. This is because SenSeNet can concentrate on these significant sentences, which is critical in the KG task. \textit{\textbf{Finally}}, when introducing the reinforcement learning, SenSeNet still has improvements compared to the baselines, which proves the generality of our model. It further demonstrates that our model is also a supplement to reinforcement learning method.

The results of absent keyphrase are shown in Table \ref{tab:absent}. We observe that our model gets the best result in all datasets on both $F_{1}@5$ and $F_{1}@M$ regardless of using reinforcement learning. According to the experimental results, our model achieves strong performance compared to other baseline models by a large margin on absent keyphrase, for each metric increases by 10$\%$ approximately. Because our model focuses on the sentences which play an important role in entire abstract. Besides, although we cannot extract keyphrases directly from some significant sentences, they usually contain central themes of a document and guides the model to generate more accurate absent keyphrases.

\subsection{Semi-Present Keyphrase Prediction}
To further explore the reason why our model performs better on generating absent keyphrases, we compare the performance of our model on semi-present keyphrase with all baseline models. 
Due the semi-present keyphrases are the subset of absent keyphrases, and according to statistics, semi-present keyphrase accounts for $19.1\%$ of absent keyphrase.
Therefore, it is very important to improve the accuracy on semi-present.

In Table \ref{tab:semi-present}, we compare the performance of our model on semi-present keyphrase with all baseline models. We select the number of keyphrase correctly predicted and use recall as the evaluation metric. We evaluate models' performance on semi-present and absent w/o(without semi-present) keyphrase. The results demonstrate that our model can predict semi-present and absent w/o keyphrase more accurately, especially semi-present keyphrase. It shows that modeling the logical document structure can greatly improve the performance of models on generating semi-present keywords.

\subsection{Keyphrase Prediction with Various Sentence Number}
\begin{figure}[t]
\centering
\includegraphics[width=1.0\linewidth]{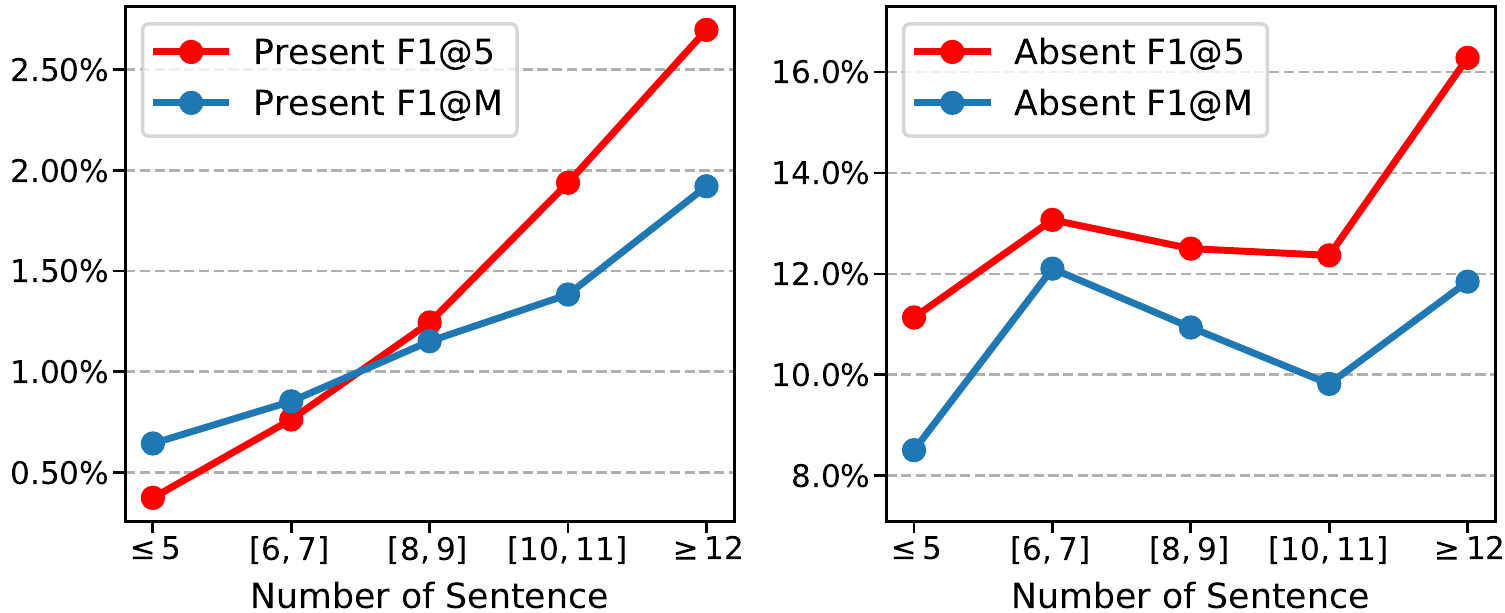}
\caption{Present keyphrase and Absent keyphrase relative gain ratios(F1@5 and F1@M measure) on various number of sentences.} 
\label{fig:SentenceVsGain}
\end{figure}

To investigate our model's capability of generating keyphrases on the long documents, we divide test dataset into 5 parts based on the number of sentence, and we reveal the relative improvement gain in $F_{1}@5$ and $F_{1}@M$ respectively on both present and absent keyphrase. The results are shown in Figure \ref{fig:SentenceVsGain}.
Both subfigures show that the rate of improvement of our model relative to baseline increases with the number of sentences growth on generating present and absent keyphrase respectively. It fully demonstrates that selecting significant sentences can greatly reduce irrelevant information, and thus improve the performance of model.`

In addition, we count the average sentence number and average significant sentence number(the sentence which includes present and semi-present keyphrase) in the datasets. The results show that there are about 7.6 sentences in a case, but only 3.9 sentences are significant sentence, accounting for 53.4\%, which means lots of sentences are irrelevant to the topics in the document. Therefore, selecting significant sentences is benefit for generating keyphrase.

\begin{figure*}[!ht]
\centering
\includegraphics[width=.95\linewidth]{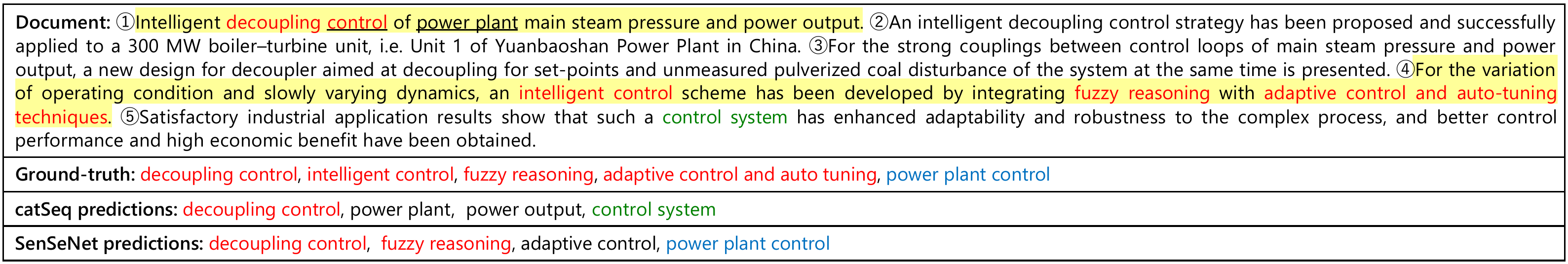}
\caption{Case study for catSeq(baseline) and SenSeNet(our model). The yellow part are significant sentences which is our model considered. The red words represent the present keyphrases, the blue words represent the absent keyphrase (also semi-present kp in this case). The green words represent the wrong keyphrase but exist in document. The underlined words represent the source of semi-present keyphrase. } 
\label{fig:case}
\end{figure*}

\begin{figure*}[!t]
\centering
\includegraphics[width=0.95\linewidth ]{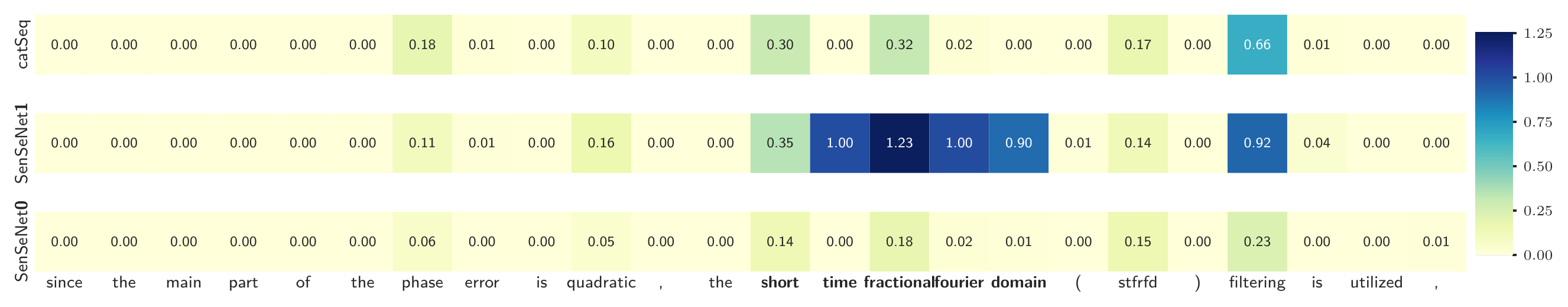}
\caption{The visualization of the probability of words predicted in three different states: baseline without SenSeNet, model with the embedding information of significant or irrelevant. The bold words represent the ground-truth keyphrase.}
\label{fig:Visualization}
\end{figure*}

\subsection{Case Study}
As shown in the Figure \ref{fig:case}, we compare the predictions generated by catSeq and SenSeNet on an article sample. Our model selects two significant sentences about title (sentence 1) and methodology (sentence 4) respectively, and almost all keyphrases are included in these two sentences. By paying attention to sentences with fewer noises, our model successfully predicts a present keyphrase and a semi-present keyphrase more than catSeq. CatSeq generates \textit{control system} which is a wrong prediction, but our model does not generate it because this phrase exists in the irrelevant sentence. Our model can reduce the probability of generating the wrong keyphrase in irrelevant sentence.  
\subsection{Visualization}
It is enlightening to analyze the influence of choosing important sentences on the decision-making of the keyphrase generation models. In Figure \ref{fig:Visualization}, three heat maps represent the sum of attention in different time steps on basic catSeq model, while determining the sentence is ``significant'' or ``irrelevant'' in SenSeNet. 

When in the basic catSeq model (first line), the probability of predicting each word is low, except for words like \textit{filtering}. Then when determining  the sentence is important(second line), the probability of some words of this sentence  have risen so much such as \textit{time}, \textit{fractional}, \textit{fourier} and \textit{domain} which is a keyphrase. It fully explains that selecting significant sentences can make the probability of words in sentences increased. And when determining the sentence is irrelevant (last line), the probability of all words reduces a lot even if the words have a high probability ever. It proves that the significance information can greatly affect the decision of the original model.
 
\subsection{Generalization}
To investigate the generalization of our method, we compare the result of keyphrase prediction with or without SenSeNet under different frameworks. As shown in table \ref{tab:Generalization},
the performance of the models are improved consistently regardless of which framework we used or whether reinforcement learning is added. It means that the model we proposed has a strong generality.

We analyze the reason from two perspectives: (1) our model has an independent significance module for introducing document structure. The functions of the module include collecting sentence's information, getting the sentence-level representation, and give the result of the significance judgment. (2) We use the embedding of "1" or "0" as the meta-sentence inductive bias, and add it to hidden state rather than directly using the binary gate to completely delete irrelevant information, which avoid the loss of the original text information. As a result, our model can be applied to any framework.

\begin{table}[tp]  
	\centering \small
	\scalebox{0.8}{
	\begin{tabular}{l|c|c|c|c}
        \toprule
		\multirow{2}{*}{\textbf{Model}} & \multicolumn{2}{c|}{\textbf{Present}} &
		\multicolumn{2}{c}{\textbf{Absent}}
		\\ 
		& \multicolumn{1}{c}{w/o SSN} &  \multicolumn{1}{c|}{w/ SSN} & \multicolumn{1}{c}{w/o SSN} &  \multicolumn{1}{c}{w/ SSN}  \\
        \midrule
		TF & 0.273 & 0.278 & 0.015 & 0.016 \\
		LSTM  & 0.289 & 0.293 & 0.015 & 0.016 \\
		GRU  & 0.291 & 0.296 & 0.015 & 0.017 \\
		TF+RL & 0.300 & 0.306 & 0.020 & 0.026 \\
		LSTM+RL & 0.310 & 0.316 & 0.024 & 0.029 \\
		GRU+RL  & 0.310 & 0.320 & 0.024 & 0.030 \\
        \bottomrule
	\end{tabular}
	}
	\caption{$F_{1}@5$ result of keyphrase prediction with or without SenSeNet in the different encoder-decoder architectures on KP20k. ``TF'' indicates Transformer.}
	\label{tab:Generalization}
\end{table}



\section{Conclusion}
In this paper, we propose a novel method named SenSeNet for keyphrase generation, which automatically  estimate whether the sentences are tended to generate the keyphrase. We use a straight-through estimator to solve the model discontinuity problem. We incorporate a weakly-supervised signal to guide the selection of significant sentences efficiently. The experiment results show that our model can successfully generate the present keyphrase and absent keyphrase. In addition, our model and the training method can be applied to most encoder-decoder architectures. Further analysis suggests that our model has an edge on semi-present keyphrase, although predicting absent keyphrase is challenging.

\bibliography{aaai21}

\end{document}